# Right Answer, Wrong Reason: Accuracy, Consistency, and Consensus Are Misleading Indicators of LLM Faithfulness in Clinical Decision Support

Bharath Kumar Bolla
Institute of Product Leadership
Bengaluru, India
bolla111@gmail.com

Chaitali Deb Purkayastha
Independent Researcher
Bengaluru, India
chaitalidebp@gmail.com

Vishnu Surya Reddy Nandi
Amazon, Inc
Seattle, USA
vishnusurya11@gmail.com

**Abstract— Clinical Large Language Models (LLMs) achieve strong medical-exam accuracy; however, a correct answer does not guarantee that the explanation names the concepts that actually drove the decision. We introduce three lightweight, directly interpretable metrics for this faithfulness gap: the Explanation Stability Index (ESI), which measures reasoning consistency across repeated queries; the Causal Faithfulness Score (CFS), which tests whether cited concepts drive predictions via concept ablation; and the Perturbation Stability Score (PSS), which measures robustness to semantic-preserving paraphrases. By evaluating six LLMs on 150 MedQA-USMLE questions (900 model-question observations), we found that only 23.3% of the cited clinical concepts were causally necessary. Correct answers had lower CFS than incorrect answers (0.212 vs. 0.398), answer consistency negatively predicted CFS (Spearman r = -0.466), and model pairs could agree on answers while sharing only 8.8% of cited reasoning concepts. These results show that accuracy, consistency, and consensus are incomplete safety signals for clinical decision-making support. The evidence is behavioral rather than mechanistic: concept ablation tests counterfactual sensitivity of outputs, not internal circuits.**

## I. INTRODUCTION

Clinical LLMs now show strong performance on standardized medical examinations and clinical QA benchmarks [1]–[3]. However, high accuracy alone does not make a system safe for clinical decision-making support. A clinician reviewing an AI recommendation must trust not only the answer but also the reasoning because a plausible explanation can differ from a faithful explanation [4], [5]. This distinction is important in medicine: a fluent but causally empty explanation may encourage unsafe generalizations in novel cases.

Consider a medical student who memorizes "elderly patients with sudden confusion usually have delirium." They may score well on multiple-choice exams. However, if the correct answer hinged on a subtle ECG finding, the student's reasoning was fundamentally wrong and would fail when the pattern changed. The same risk applies to clinical LLMs that cite plausible-sounding but causally irrelevant concepts.

Rationale evaluation work motivates checking whether the cited evidence actually supports a prediction [6]. Chain-of-thought prompting [7] improves LLM performance, but these explanations can be systematically unfaithful [8], [9] because models may construct post hoc narratives to justify a predetermined answer. In the clinical domain, this creates a faithfulness gap between what a model says and how it behaves to reach a decision. Benchmarks focused on accuracy overlook the three distinct properties required for trustworthy explanations: stability, causal faithfulness, and robustness to harmless rephrasing.

This study makes five contributions. First, we introduce ESI, CFS, and PSS as lightweight, directly interpretable faithfulness metrics that do not require an auxiliary model for evaluation. Second, we report a cross-model clinical evaluation of 150 MedQA-USMLE questions. Third, we identify the correctness paradox: correct answers often have lower causal faithfulness than incorrect answers. Fourth, we show that answer consistency and consensus can mask divergent or post-hoc reasoning. Fifth, we provided category-level faithfulness scores that indicate which types of clinical concepts require the most scrutiny.

## II. RELATED WORK

Faithfulness in NLP explanations was formalized by Jacovi and Goldberg [4], who distinguished faithful explanations from plausible ones. Lyu et al. [5] survey methods for evaluating faithfulness and identify perturbation-based evaluation as a practical approach for black-box models. ERASER [6] similarly frames the evaluation of rationale around whether the selected evidence supports the predictions. Our work applies this line of evaluation to clinical decision support, where the cost of an unfaithful explanation is not only an interpretability error but also a possible clinical over-trust.

Wei et al. [7] showed that chain-of-thought prompting improves reasoning performance, but Turpin et al. [8] and Lanham et al. [9] show that chain-of-thought explanations can misrepresent why a model selected an answer. These studies motivate evaluating explanation faithfulness separately from final answer accuracy. We extend this concern to clinical QA, where a model may answer correctly while citing concepts that do not govern the model's behavioral response.

The closest prior work to our CFS is that of Matton et al. [10], who introduced causal concept faithfulness using an auxiliary LLM to generate realistic counterfactuals and a Bayesian hierarchical model to quantify the concept-level effects. Gat et al. [11] also used LLM-generated counterfactuals for faithful black-box NLP explanations. Our CFS differs in three ways: it uses direct text ablation rather than LLM-generated counterfactuals, uses a binary answer-change criterion that is directly interpretable, and supports cross-model clinical comparisons without additional inference infrastructure.

Self-consistency decoding treats the agreement among sampled answers as a quality signal [12]. For explanation stability, we used the mean pairwise SBERT cosine similarity [13], whereas Becker and Soatto [14] measured confidence through stable explanations. Our behavioral metrics complement calibration and feature-attribution methods [15]-[17] but do not claim to identify internal mechanisms. This distinction addresses an important limitation: CFS is a black-box counterfactual sensitivity test, not a mechanistic interpretability method.

## III. METHODOLOGY

### *A. Dataset and Models*

We used 150 questions from MedQA-USMLE [3], a benchmark of physician-level clinical reasoning across diverse specialties, including internal medicine, surgery, pharmacology, and pediatrics. The questions were randomly sampled to preserve the original difficulty distribution. Each question presented a clinical vignette with five answer options and required multistep reasoning. Each model received a structured prompt eliciting step-by-step REASONING, KEY_FACTORS (3-5 clinical concepts), a single ANSWER letter, and a CONFIDENCE score (0-100).

Six models were selected to span the five axes of variation relevant to real-world clinical deployment. (1) Accessibility: GPT-4o-mini (OpenAI) represents widely deployed commercial APIs, while the remaining five are open-weight models. (2) Domain specialization: MedGemma-27B (Google) is fine-tuned on medical literature [5]; OpenBioLLM-8B is trained exclusively on biomedical corpora; GPT-4o-mini, GLM-4.7-Flash, DeepSeek-R1, and MiniMax-M2.5 are general-purpose. This contrast directly tests whether domain tuning improves the explanation faithfulness. (3) Architecture: DeepSeek-R1-Distill-Qwen-32B is a reasoning-specialised model distilled from a chain-of-thought teacher, making it the natural candidate for highest CFS; MiniMax-M2.5 uses a mixture-of-experts architecture. (4) Scale: models range from 8B parameters (OpenBioLLM) to 27–32B (MedGemma, DeepSeek), with commercial models at undisclosed scale, allowing us to assess whether model size correlates with faithfulness. (5) Organizational diversity: US commercial (OpenAI, Google), Chinese open-weight (Zhipu AI, DeepSeek, MiniMax), and community biomedical (OpenBioLLM) models are represented, guarding against findings being an artifact of any single organization's training approach. The observed accuracies ranged from 0.550 (OpenBioLLM) to 0.871 (MedGemma), confirming that the selected models covered the full range of performance.

### *B. Explanation Stability Index (ESI)*

Each question was presented 10 times independently at a temperature of 0.7. ESI is the mean pairwise cosine similarity of Sentence-BERT [13] embeddings of the REASONING field across all C(10,2) = 45 pairs of sentences. Equation (1) is as follows:

$$ESI(q,m) = (2/n(n-1)) \Sigma_{i<j} \cos(e_i, e_j) \quad (1)$$

We additionally computed answer consistency (AC = the fraction of runs that give the same answer), key-factor Jaccard (KFJ = the mean Jaccard of concept sets across run pairs), and confidence spread. Unlike Becker and Soatto [14], who used stability to estimate confidence, we treat stability as a distinct property and test whether it aligns with causal faithfulness.

### C. Causal Faithfulness Score (CFS)

For each model-question pair, the model first answered the full question and cited five key clinical concepts. Each concept was then ablated by removing its explicit mention from the question text, and the model was re-queried. A concept c is considered genuinely causal if ablation changes the answer. This is a behavioral test: ablation may remove contextual coherence or leave parametric knowledge intact, so CFS should be interpreted as output sensitivity rather than proof of an internal causal mechanism. Equation (2) enumerates the CFS.

$$CFS(q,m) = |\{c \in C : A(q \setminus c) \neq A(q)\}| \,/\, |C| \quad (2)$$

| Box 1: CFS Worked Example (medqa_1030, GPT-4o-mini) |
|---|
| Case: A 3466-g female newborn (38 wks), micrognathia, cleft palate, |
| single, overriding, great vessel. Mother: lithium exposure (5 wks), |
| alcohol use disorder (now sober). Most likely evaluation finding |
| Options: A) Low PTH  B) Double bubble sign  C) Chr-5p del |
| D) Tricuspid elongation  E) Bilateral cataracts |
| |
| **Step 1 - Initial model response: Answer = A (DiGeorge syndrome)** |
| Cited concepts: [1] Congenital Dysmorphic Features |
| [2] Cardiac Anomalies |
| [3] Maternal Substance Use |
| [4] DiGeorge Syndrome (22q11.2 deletion) |
| [5] Hypoparathyroidism |
| |
| **Step 2 - Concept ablation [1] "Congenital Dysmorphic Features":** |
| The question omits micrognathia/cleft palate references. |
| Re-query -> Answer = A (unchanged) => [1] = cited-but-unused |
| |
| **Step 3 - Ablate concept [2] "Cardiac Anomalies":** |
| The question omits the overriding great vessel reference. |
| Re-query -> Answer = B (changed)   => [2] = genuinely causal |
| |
| Similarly: [3] unused \| [4] causal \| [5] causal |
| **CFS = 3/5 = 0.60  (concepts [1] and [3] are post-hoc rationalisations)** |

### *D. Perturbation Stability Score (PSS)*

Three semantically equivalent paraphrases are generated per question using GPT-4o-mini with the prompt: "Rewrite this clinical question maintaining all clinical details, answer options, and the correct answer, while altering the phrasing and sentence structure." Answer stability is the fraction of paraphrases that elicit the same answer as the original. Because GPT-4o-mini is also an evaluated model, PSS results involving it are interpreted with caution; the paraphraser may introduce a model-specific style bias.

| Box 2: PSS Worked Example (medqa_1030, GPT-4o-mini) |
|---|
| Original question: "A 3466-g female newborn... micrognathia, cleft |
| palate, and a single overriding great vessel..." |
| **Original answer: A \| Confidence: 85** |
| |
| Paraphrase 1: "A female infant weighing 3466 g was born at 38 weeks |
| gestation... distinct facial features and cardiac single overriding vessel..." |
| **Answer: C (Chromosome 5p deletion) <- CHANGED** |
| |
| Paraphrase 2: "A 7 lb 10 oz female born at 38 weeks... dysmorphic |
| features including cleft palate... overriding great vessel..." |
| Answer: A (unchanged) |
| |
| Paraphrase 3: "A female newborn (38 wks)... micrognathia, cleft palate, |
| low-set ears, overriding great vessel... lithium exposure history..." |
| Answer: A (unchanged) |
| |
| **Answer stability = 2/3 = 0.667 \| Concept Jaccard = 0.042 \| Conf. drift = 5.0** |
| Interpretation: identical clinical content, different surface phrasing, |
| yet the model changes the diagnosis on one of the three paraphrases. |

### *E. Extended Analyses (Experiments 8, 9, 11, 12, 13)*

Five zero-cost analyses were performed on the aggregated CSVs. Exp. 8 compares CFS for correct vs. incorrect answers (Mann-Whitney U). Exp. Nine tests of the AC vs. CFS correlation (Spearman; Kruskal-Wallis across quartiles). Exp. The 11 scores six clinical concept categories for the faithfulness fraction. Exp. Table 12 compares the pairwise answer agreement and concept agreement across the 15 model pairs. Exp. 13 stratifies questions into EASY/MEDIUM/HARD by cross-model answer success (EASY n = 102, MEDIUM n = 25, HARD n = 23). Because the HARD contains only 23 questions, difficulty-stratified claims are exploratory. All means carried 95% bootstrap CIs (10,000 resamples, seed = 42).

## IV. RESULTS

### *A. Core Faithfulness Metrics*

Table I summarizes the three core metrics. A mean CFS of 0.233 means fewer than 1.2 of 5 cited clinical concepts are genuinely causally related. This does not prove that the model lacks internal use of the concept, but it does show that the surface explanation often fails a direct behavioral counterfactual test. The Key Factor Jaccard ranged from 0.027 (GLM-4.7) to 0.143 (GPT-4o-mini), indicating that the models frequently generated different cited concepts across repeated runs.

TABLE I: CORE METRICS FOR EACH MODEL. ESI = Explanation Stability Index; AC = Answer Consistency; KFJ = Key Factor Jaccard; CFS = Causal Faithfulness Score; Ans. Stab. = Answer Stability under paraphrasing.

| Model | Acc | ESI | AC | KFJ | CFS | Ans. Stab. |
|---|---|---|---|---|---|---|
| **GPT-4o-mini** | 0.79 | 0.89 | 0.92 | 0.13 | 0.25 | 0.84 |
| **GLM-4.7** | 0.72 | 0.83 | 0.83 | 0.03 | 0.28 | 0.70 |
| **DeepSeek-R1** | 0.82 | 0.66 | 0.92 | 0.11 | 0.33 | 0.86 |
| **MiniMax** | 0.83 | 0.88 | 0.91 | 0.06 | 0.13 | 0.86 |
| **MedGemma** | 0.87 | 0.84 | 0.96 | 0.07 | 0.23 | 0.90 |
| **OpenBioLLM** | 0.55 | 0.79 | 0.73 | 0.03 | 0.28 | 0.49 |

*B. Correctness Paradox (Exp. 8)*

Table II and Fig. 1 show CFS stratified by answer correctness. Correct answers had CFS = 0.212 [0.194, 0.230], whereas incorrect answers had CFS = 0.398 [0.358, 0.439]. Spearman (correct_rate, CFS) = -0.44 (Pearson = -0.38; both $p < 0.001$). This does not imply that incorrect answers are preferable; rather, it shows that benchmark-correct answers can arise from shortcuts that are later rationalized with plausible clinical language.

TABLE II: CFS BY ANSWER CORRECTNESS (Exp. 8). Mean [95% CI]; p-value from two-sided Mann-Whitney U test.

| Model | N corr. | CFS correct | CFS incorr. | p-val | Sig. |
|---|---|---|---|---|---|
| **GPT-4o-mini** | 124 | 0.23 [0.19, 0.27] | 0.35 [0.25, 0.45] | 0.02 | * |
| **GLM-4.7** | 114 | 0.20 [0.16, 0.25] | 0.50 [0.41, 0.58] | <0.01 | *** |
| **DeepSeek-R1** | 128 | 0.30 [0.26, 0.34] | 0.52 [0.40, 0.64] | <0.01 | *** |
| **MiniMax** | 128 | 0.09 [0.05, 0.12] | 0.41 [0.30, 0.53] | <0.01 | *** |
| **MedGemma** | 133 | 0.21 [0.18, 0.25] | 0.36 [0.25, 0.49] | 0.01 | * |
| **OpenBioLLM** | 89 | 0.25 [0.19, 0.30] | 0.32 [0.25, 0.39] | 0.10 | ns |

*MedGemma and OpenBioLLM: ns due to small n_incorrect. All others: p < 0.05 or p < 0.001.*

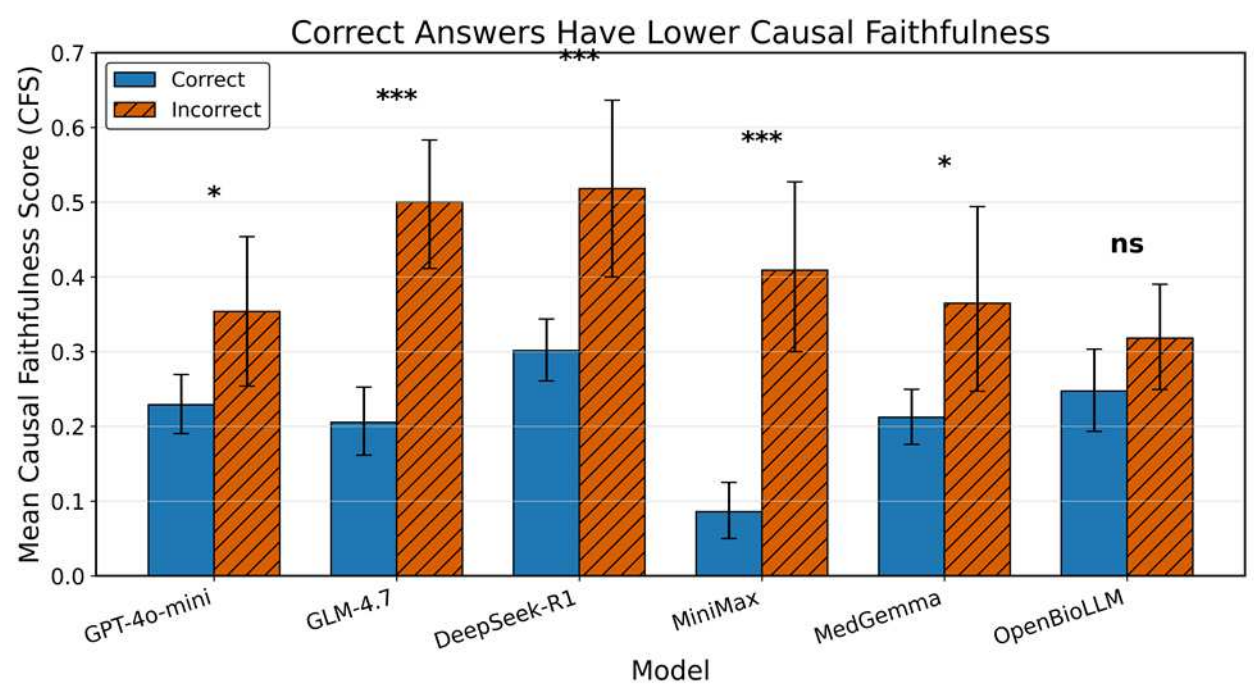

Fig. 1. CFS for correct (solid) and incorrect (hatched) answers per model. Error bars indicate 95% bootstrap CI. Stars: Mann-Whitney U significance.

*C. Self-Consistency Is a Misleading Proxy (Exp. 9)*

Spearman(AC, CFS) = -0.466 (Pearson = -0.481; both $p < 0.001$, $n = 900$). This challenges the assumption behind self-consistency decoding [12], which states that higher answer agreement is sufficient evidence of greater reliability. This relationship holds across all six models (Spearman r ranging from -0.32 to -0.65). Kruskal-Wallis across AC quartiles: $H = 195.2$, $p < 0.001$. The MedGemma Q1 CFS was 0.900, while that of MiniMax Q4 CFS = 0.043. A model that repeatedly provides the same answer may have a stable shortcut rather than stable reasoning.

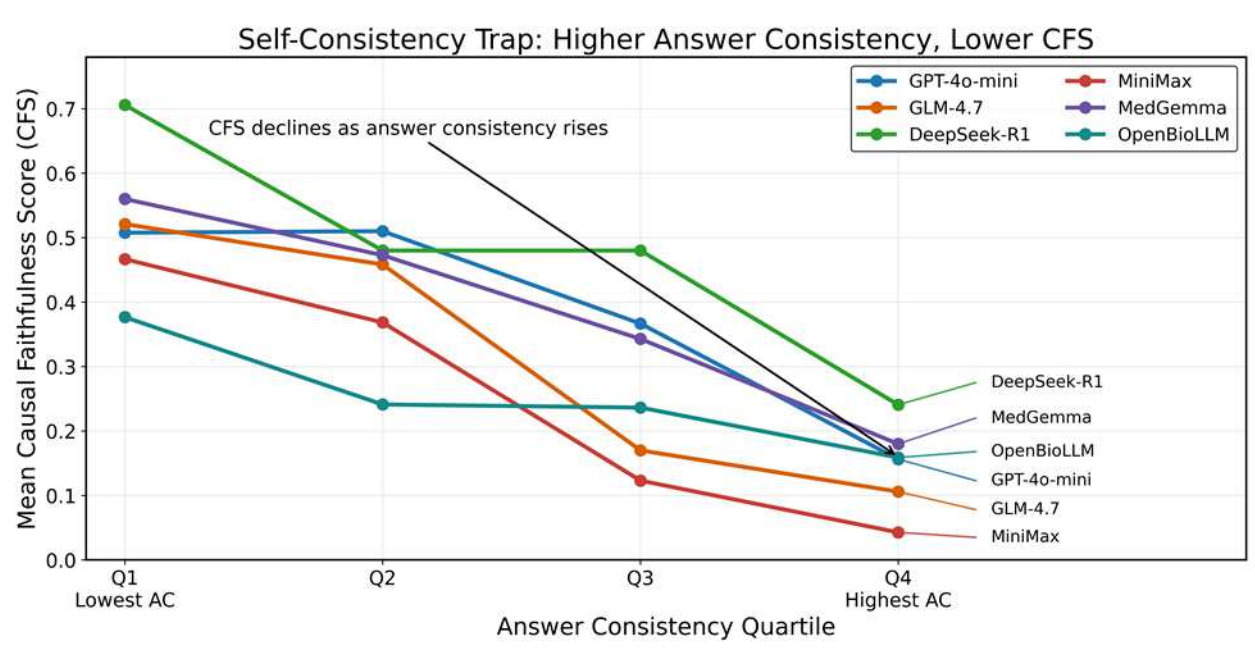

Fig. 2. Mean CFS by answer consistency quartile per model. Q1: AC < 0.70; Q4: AC >= 0.95. All six models showed that CFS declined monotonically as confidence increased.

*D. Model consensus mask rationalization (Exp. 12)*

Across all 15 model pairs, the mean answer agreement was 0.812, while the mean concept-level Jaccard was only 0.088, with a rationalization gap of 0.724. MiniMax and MedGemma agreed on the answer 94% of the time, yet shared only 9.3% of the cited concepts. Table III and Fig. 3 show the largest gaps between the two. Answer-level consensus is therefore an incomplete reliability signal unless the models also converge on the clinical concepts that support the answer.

TABLE III: TOP-5 RATIONALIZATION GAPS (Exp. 12). Ans. Agr. = same answer fraction; Concept Agr. = mean Jaccard.

| Model Pair | Ans. Agr. | Concept Agr. | Gap |
|---|---|---|---|
| **MiniMax / MedGemma** | 0.94 | 0.09 | 0.85 |
| **GPT-4o-mini / MedGemma** | 0.91 | 0.07 | 0.84 |
| **DeepSeek-R1 / MedGemma** | 0.89 | 0.08 | 0.81 |
| **GPT-4o-mini / MiniMax** | 0.89 | 0.12 | 0.77 |
| **DeepSeek-R1 / MiniMax** | 0.87 | 0.12 | 0.75 |

*Causal-only Jaccard (genuinely causal concepts) averages 0.042 across all pairs, approaching zero.*

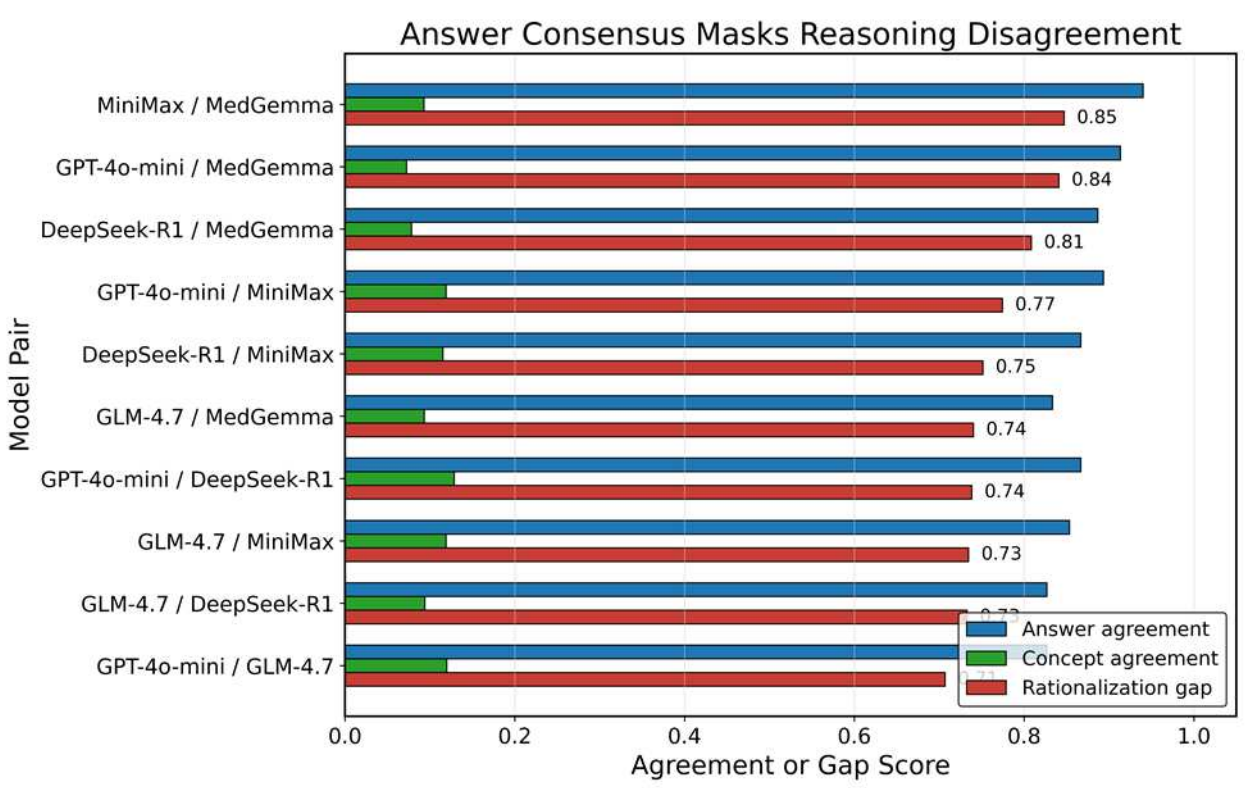

Fig. 3. Rationalization gap for top-10 model pairs. Blue = answer agreement; green = concept agreement; red = gap. All pairs exceeded a gap of 0.70.

### *E. Faithfulness and Difficulty (Exp. 13)*

The questions were stratified as EASY (n = 102), MEDIUM (n = 25), and HARD (n = 23) using cross-model answer success. Spearman (difficulty, CFS) = +0.408 ($p < 0.001$); Kruskal-Wallis CFS H = 131.3, $p < 0.001$. For easy questions, models appeared to rely more on shortcuts (MiniMax EASY CFS = 0.035); for hard questions, they showed higher CFS (MiniMax HARD = 0.382; DeepSeek HARD = 0.582). Because HARD has only 23 questions, we describe this as an exploratory signal rather than evidence that correct hard answers are lucky guesses. Figs. 4 and 5 visualize the pattern.

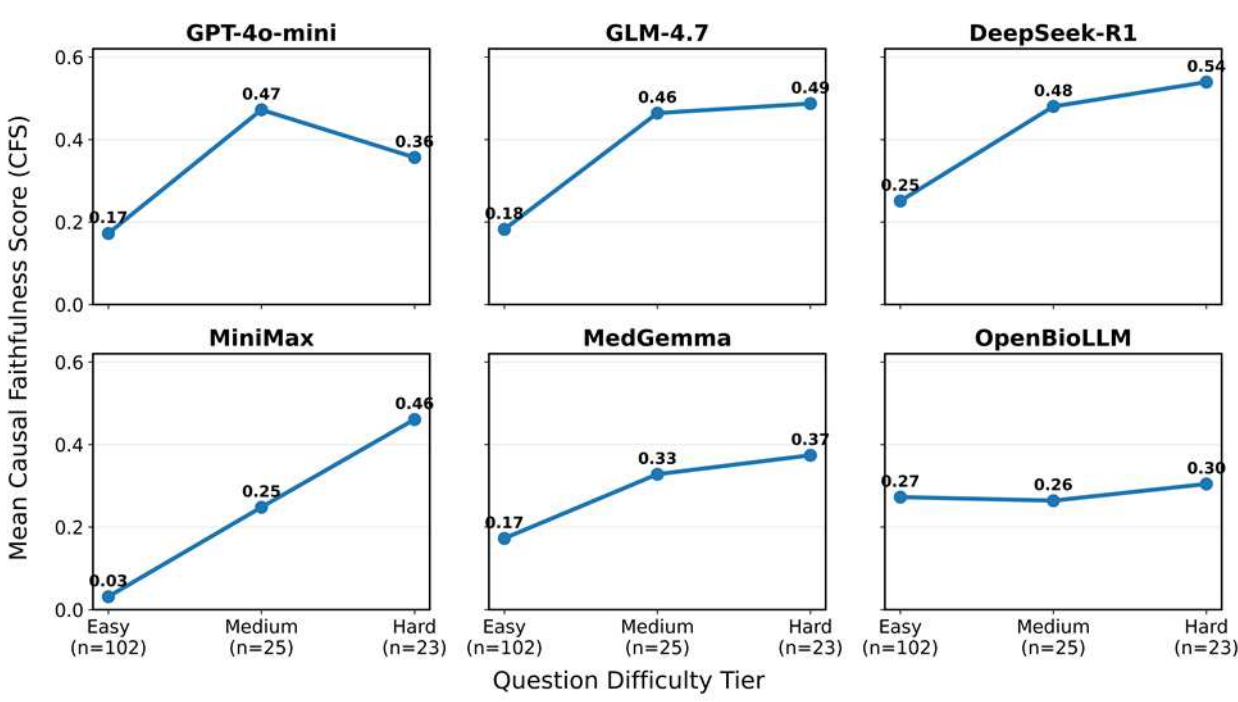

Fig. 4. CFS vs. difficulty per model (2 × 3 grid). Shading: green=EASY, orange=MEDIUM, red=HARD. The positive slope in all six panels confirms that harder questions elicit more causally grounded reasoning.

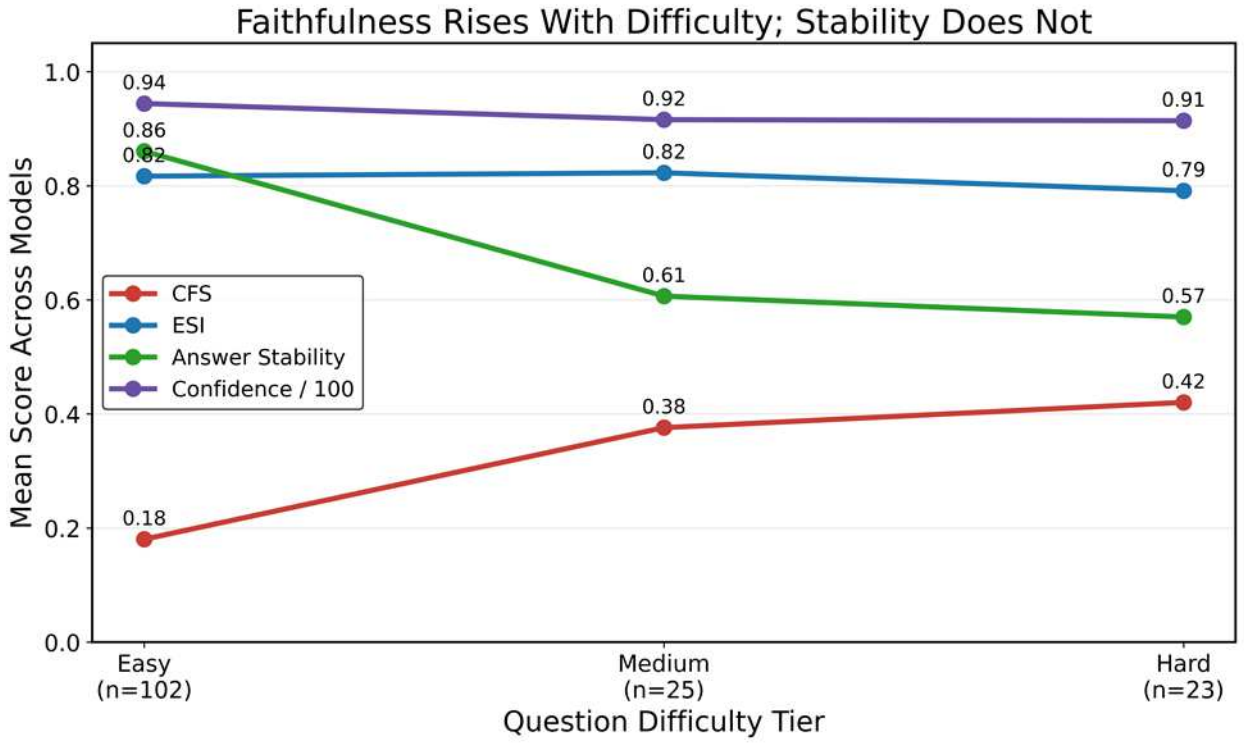

Fig. 5. ESI, CFS, answer stability, and confidence by difficulty tier. The CFS rises with difficulty, whereas the ESI remains comparatively flat.

### *F. Category-Level Faithfulness (Exp. 11)*

Fig. 6 shows faithfulness by clinical concept category. SYMPTOM concepts were the least faithful: only 19.2% genuinely drove the answer (n = 156). ANATOMY was highest at 31.8% (n = 170). DIAGNOSTIC (26.1%), PATHOLOGY (21.2%), and TREATMENT (31.1%) fell between. Returning to Box 1, the model cited "Congenital Dysmorphic Features" (SYMPTOM) and "Maternal Substance Use" (OTHER), both of which were decorative, while "Cardiac Anomalies" (ANATOMY), "DiGeorge Syndrome" (PATHOLOGY), and "Hypoparathyroidism" (PATHOLOGY) were the causal terms. Therefore, symptom-based AI explanations should be treated with caution.

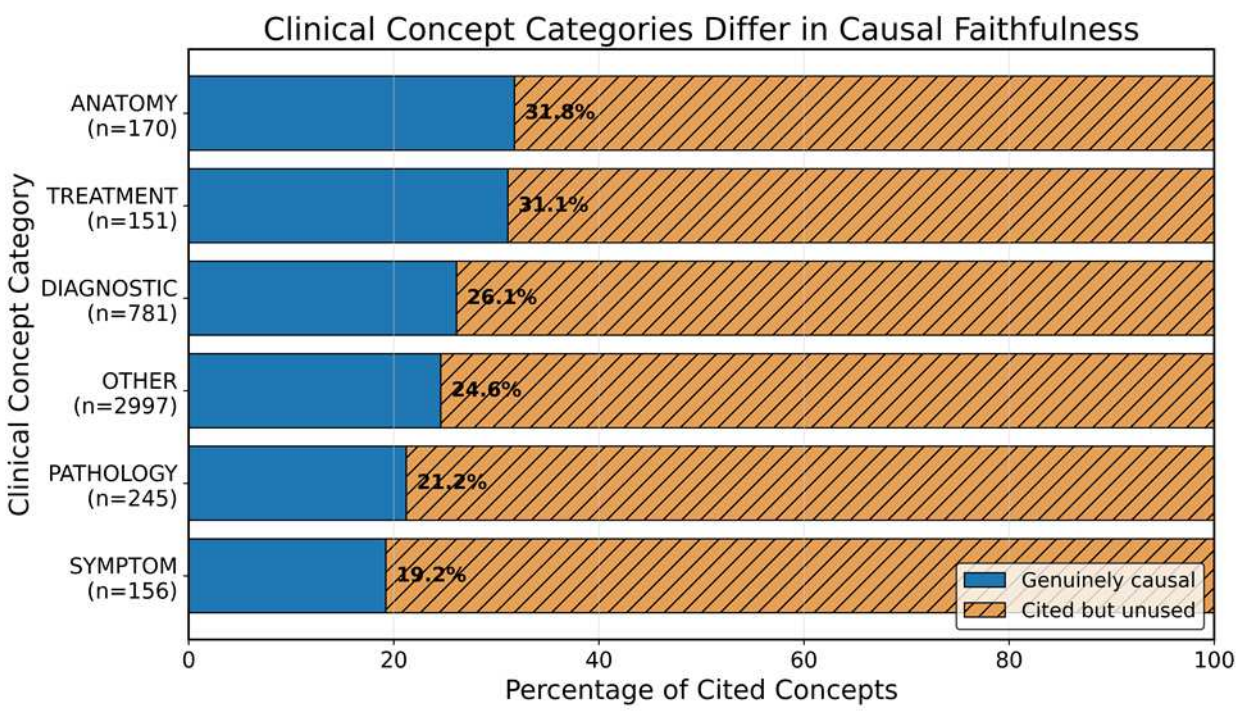

Fig. 6. Category faithfulness: genuinely causal (solid) vs. cited-but-unused (hatched) by clinical concept type. SYMPTOM concepts were the least faithful, whereas ANATOMY concepts were the most faithful among the reported categories (31.8% causal).

# V. DISCUSSION

## *A. The Faithfulness Gap: A Unified View*

All five experiments point to the same concern: clinical AI models can obtain the correct answer while citing weakly causal reasons. Therefore, their explanations should be treated as hypotheses about reasoning rather than evidence of reasoning. The usual reliability signals--accuracy, consistency, stability, and fluency--do not guarantee causal faithfulness. This pattern is not a failure of any single model but a recurring property across all the systems tested here.

The three metrics are complementary and non-redundant. The ESI measures surface consistency: does the explanation look similar across runs? CFS measures causal grounding: do cited reasons drive the answer under ablation? PSS measures robustness: Does rephrasing preserve the answer and explanation? No model excels in all three. DeepSeek leads on CFS but has a lower ESI; MedGemma leads on PSS but shows moderate CFS; MiniMax shows high consistency but low faithfulness. This non-redundancy enables the evaluation of all three dimensions.

## *B. Clinical Safety Implications*

The correctness paradox (Exp. 8) means that a correct AI answer is not a certificate of trustworthy reasoning. A model can learn a high-performing clinical shortcut and then produce a plausible narrative after making a decision. In clinical use, this is most important for rare presentations, comorbidities, and edge cases that differ from benchmark distributions.

The self-consistency trap (Exp. 9) is also important. A model that always provides the same answer may have a stable shortcut rather than a stable reasoning. Using answer consistency alone as a deployment filter could select models that are confident and repeatable but weakly so. The category findings (Exp. (11) provide practical guidance: when an AI explanation relies primarily on symptoms, clinicians should verify whether those symptoms actually determine the answer.

The consensus illusion (Exp. (12) cautioned against answer-only multi-model verification. Two models agreeing on an answer while sharing little concept-level reasoning do not necessarily corroborate each other. More useful ensemble verification would require agreement on both the answer and the clinical concepts that support it.

## *C. Model-Specific Clinical Recommendations*

Table IV maps each model to its primary failure mode and monitoring priorities. These recommendations are not deployment-approval documents. MedGemma and DeepSeek-R1 are the closest in this benchmark: MedGemma has the best paraphrase robustness (AS = 0.907), whereas DeepSeek-R1 has the highest overall CFS (0.324). Both still require human oversight of the cited concepts. Broader clinical safety work on hallucinations, healthcare AI evaluation, regulation, and the limitations of post-hoc explainability supports treating these metrics as screening tools rather than certification [18]-[22].

*TABLE IV: MODEL-SPECIFIC DEPLOYMENT RECOMMENDATIONS. This is based on the evidence from Experiments 1, 3, 8, 9, 11, 12, and 13.*

| Model | Strength | Critical Weakness | Recommendation |
|---|---|---|---|
| **GPT-4o-mini** | Best KFJ (0.14); balanced across all metrics | CFS low (0.23); shortcuts dominate on easy questions | Use with concept-level human verification |
| **GLM-4.7** | High accuracy | Strongest AC-CFS coupling (r=-0.62); most misleading when confident | Do not use AC as a quality proxy here |
| **DeepSeek-R1** | Highest CFS (0.32); most causally grounded | Lowest ESI (0.59); explanation text unstable across runs | Strongest CFS in this benchmark; verify reasoning each call |
| **MiniMax** | Highest AC (0.93); consistent answers | Q4 CFS = 0.05; nearly all reasons decorative when most confident | Avoid for tasks requiring trustworthy explanations without added checks |
| **MedGemma** | Best answer | Faithfulness collapses | Best paraphrase stability in this |

|  | stability (0.91); Q1 CFS = 0.90 | in high-confidence regime | benchmark; verify cited concepts |
|---|---|---|---|
| **OpenBioLLM** | Open-source; accessible | Worst answer stability (0.48); changes diagnosis on rephrasing | Requires substantial validation before clinical use |

None of the models were suitable for unsupervised clinical deployment based on these results. MedGemma and DeepSeek-R1 are the strongest candidates for supervised workflows, provided that their cited concepts are independently checked.

### *D. Limitations*

This study used 150 MedQA-USMLE questions that may not represent all clinical specialties, rare diseases, longitudinal records, or real clinical notes. The ablation protocol tests counterfactual sensitivity, not mechanistic interpretability inside the model. Removing a clinical term can alter contextual coherence, leave redundant cues intact, or fail to erase parametric knowledge; therefore, CFS can under- or overestimate true causal use. The six-category concept taxonomy is rule-based and may misclassify complex multiword concepts. PSS paraphrases were generated using GPT-4o-mini, which may have introduced a model-specific style bias. The HARD difficulty stratum contained only 23 questions; therefore, the difficulty results require replication. Future work should include clinician validation, larger multi-institution datasets, human or template paraphrases, multi-concept ablations, and mitigation methods, such as abstention or calibration.

## VI. CONCLUSION

We introduced three lightweight, directly interpretable faithfulness metrics (ESI, CFS, and PSS) that support cross-model comparisons without auxiliary evaluation models. Applied to 150 MedQA-USMLE questions across six models (900 model-question observations), the findings are consistent: only 23.3% of cited clinical concepts are causally necessary; correct answers often have lower causal faithfulness than incorrect answers; self-consistency [12] is an incomplete reliability signal; model consensus can reflect shared answers without shared reasoning; and symptom-based explanations are the least trustworthy. The central lesson is cautious but important: clinical LLM explanations should be evaluated at the conceptual level before being treated as evidence for clinical decision support.